\documentclass[10pt,twocolumn,letterpaper]{article} 

\pdfoutput=1
\usepackage{avss}
\usepackage{times}
\usepackage{epsfig}
\usepackage{graphicx}
\usepackage{amsmath}
\usepackage{amssymb}
\usepackage{mathtools}
\usepackage{algorithm}
\usepackage{algpseudocode}
\usepackage{adjustbox}
\usepackage{lipsum}
\usepackage{verbatim}
\usepackage{dblfloatfix}

\usepackage[at]{easylist} 

\usepackage{xcolor}
\usepackage{soul}
\definecolor{newcolor}{rgb}{.8,.349,.1}

\usepackage[pagebackref=true,breaklinks=true,letterpaper=true,colorlinks,bookmarks=false]{hyperref}

\avssfinalcopy 

\def\avssPaperID{} 

\ifavssfinal\fi
\begin{document}

\title{Understanding the Unforeseen via the Intentional Stance}

\author{Stephanie Stacy\\
Department of Statistics\\
UCLA\\
{\tt\small stephaniestacy@g.ucla.edu}
\and
Alfredo Gabaldon, John Karigiannis, James Kubrich, Peter Tu\\
GE Research\\
{\tt\small \{firstname.lastname, tu\}@ge.com}
}

\maketitle

\begin{abstract}
\noindent We present an architecture and system for understanding novel behaviors of an
observed agent. The two main features of our approach are the adoption of 
Dennett's intentional stance and analogical reasoning as one
of the main computational mechanisms for understanding unforeseen experiences.
Our approach uses analogy with past experiences to construct hypothetical rationales
that explain the behavior of an observed agent. Moreover, we view analogies as partial; thus multiple past experiences can be blended to analogically explain an unforeseen event, leading to greater inferential flexibility. We argue that this approach results
in more meaningful explanations of observed behavior than approaches based on
surface-level comparisons. A key advantage of behavior explanation over classification is the ability to i) take appropriate responses based on reasoning 
and ii) make non-trivial predictions that allow for the verification of the
hypothesized explanation. We provide a simple use case to demonstrate novel experience understanding through analogy in a gas station environment.\footnote{This research was, in part, funded by the U.S. Government. The views and conclusions contained in this document are those of the authors and should not be interpreted as representing the official policies, either expressed or implied, of the U.S. Government. Use, duplication, or disclosure is subject to the restrictions as stated in Agreement HR00111990061 between the Government and GE.}
\end{abstract}

\vspace{-10mm}

\section{Introduction}

\noindent Over the past decade, significant progress has been made in developing computer vision systems which can detect and identify objects and events in real world video feeds. However, a computer's understanding of the environment does not exhibit the depth and nuance of that formed by even a na\"{i}ve human observer. Consider, for example, a common environment encountered often in daily life: the gas station. Video analytics systems can be installed in this environment to identify customers in the scene, critical landmarks, and even events that are relevant to the station's operations (e.g., customers departing without removing fuel nozzles). However, if the station were to function without a human attendant, computer vision systems would also need to accurately detect events that are surprising or unanticipated. These situations generally arise due to randomness in human behavior where the same underlying event may be expressed differently depending on contextual factors. The autonomous attendant of the future -- when exposed to novel and unanticipated events -- must be able to ``get the gist of things'' and respond appropriately to ensure customers are safe and operations are proceeding as expected. To establish such a capability, we argue that the autonomous attendant must understand the \emph{roles} and \emph{affordances} of entities in the environment; to do this, the \emph{intentional stance} must be taken.


\subsection{The Intentional Stance}


\noindent Under Dennett's theory of the intentional stance \cite{Dennett}, the behaviors of agents in the environment should be understood by assuming they were chosen according to their own beliefs and desires.
The theory begins by considering forms of intelligence developed over long periods of reward-based learning that may exist outside of an agent's awareness; i.e., concepts are ``trapped in connectionist (neural) meshes". 
At some point, humans developed various mind tools (such as language) that allowed for the ability to: i) extract neurally encoded concepts, ii) manipulate existing concepts to analyze novel concepts and iii) store concepts in shared natural language so that individuals may rediscover concepts rather than invent them from scratch. Once agents became capable of harvesting and subsequently storing concepts, they were then able to make use of the intentional stance with respect to the behavior of other agents. That is, they could not only classify the behavior of another agent but also discern the rationale behind such behaviors. By taking the intentional stance, we assume that the agent is rational and driven by desires and beliefs towards intentions and actions. However, the agent may not actually have such levels of intentionality. In fact, the agent may be oblivious to the rationale behind its behaviors, e.g., when behavior is a product of learning as opposed to planning. Nevertheless, we argue that an observer may infer rationales underlying an actor's behavior across multiple experiences and subsequently map those rationales to new experiences through analogical reasoning.

\subsection{Analogies Between Multiple Experiences}

\noindent Douglas Hofstadter \cite{Hofstadter} argues that people possess a ``currency of thought'' where novel situations are interpreted by constructing analogies and reasoning over them using salient contextual information. In this way, prior experiences can be used to interpret novel circumstances. In the current study, we aim to construct rationales for observed behaviors by forming partial
analogies based on multiple past experiences. By combining the hypotheses resulting from each partial analogy, a rationale for novel observed behaviors can be inferred.
To this end, the analogical reasoning process in the presented architecture evolves by
i) iterating over sets of base scenarios with rationales (or explanations), ii) constructing analogies between base and target scenarios, and iii) incrementally constructing a complete rationale for observed behaviors.
Each analogy with a base scenario may yield a partial rationale in addition to other facts about the observed events. These facts and partial rationales are then added -- as new content -- to the target scenario which allows richer analogies to be formed when considering additional bases.
The synthesis of partial analogies iterates over the entire set of prior experiences and terminates after an iteration fails to generate additional hypotheses. Since each analogy adds details to the target, the order in which base scenarios are considered may affect the result as well as execution time. We thus explore the use of heuristics that consider structural as well as surface level characteristics to choose an advantageous ordering.


As an example, consider the event sequences shown in Table \ref{tab:experiences}. From Prior Experience \#1 we can derive the following rationale: third parties may approach customers with the intention of satisfying a desire. From Prior Experience \#2 we can construct the rationale: dangerous situations may cause the customer to flee. How then should the subsequent Novel Experience be interpreted? A somewhat simplistic set of partial analogies would be of the form: 3a $\rightarrow$ 1a \& 2a, 3b $\rightarrow$ 1b, and 3c $\rightarrow$ 2c. An intelligent observer might infer from these associations that the customer fled from the stranger in the third situation because they felt threatened (as was the case in the second situation when the vehicle caught on fire). While simplistic, this example illustrates the process by which an agent, without prior knowledge of carjacking events, is able to construct an explanatory rationale which can be used to inform actions, such as notifying emergency personnel.

\begin{table}
\caption{Example experiences used to construct analogy in gas station scenario.}
\begin{center}
\begin{tabular}{c l}
    \hline
    \multicolumn{2}{c}{Prior Experience \#1} \\
    \hline
    1a & Customer begins pumping gas \\
    1b & Stranger approaches customer \\
    1c & Stranger asks customer for directions \\
    1d & Stranger leaves \\
    1e & Customer finishes pumping \\
    \hline
    \multicolumn{2}{c}{Prior Experience \#2} \\
    \hline
    2a & Customer begins pumping gas \\
    2b & Vehicle catches on fire \\
    2c & Customer runs away from vehicle \\
    \hline
    \multicolumn{2}{c}{Novel Experience} \\
    \hline
    3a & Customer begins pumping gas \\
    3b & Stranger approaches customer \\
    3c & Customer runs away from vehicle \\
    \hline
    \vspace{-10mm}
\label{tab:experiences}
\end{tabular}
\end{center}
\end{table}



With this example in mind, we propose the following architecture for synthesizing partial analogies with (past) base experiences to explain novel target scenarios. First, novel experiences are obtained by observing
an actor agent whose behavior -- in accordance with the intentional stance -- is assumed to have been motivated by some unobservable rationale.
When constructing a rationale for novel observations, the observer does not know the underlying policy of the actor nor its state of mind; they can only observe the performed sequence of actions. 
The observed sequence of actions, alongside relevant background facts about
the environment and the agent, are represented as a set of propositions encoding the
target experience. 
We then compare the target against a library of base experiences for 
which the rationales are known. This comparison is done by mapping the novel experience to past experiences via the structure-mapping engine (SME) algorithm \cite{falkenhainer1989structure}. A central tenet of the SME, and of the Structure Mapping Theory \cite{gentner1983structure} behind it, is its emphasis that analogies are driven by structural (rather than surface-level) attribute similarities which lies in contrast to purely associative learning approaches. SMEs have been traditionally used to map knowledge from one domain to another; however, here we extend them to map behaviors and infer rationales.

\section{Technical Description}

\noindent Our work on behavior understanding under the intentional stance begins with
development of a simulation system we call the {\it Artificial Intentionality Engine}
(AIE). The purpose of the AIE is to establish an environment where agents (over many iterations) learn to perform actions to maximize reward in the absence of known rationales. This provides a source of observed behaviors which we term \emph{chronologies}. A description of the AIE is given in Section \ref{sec:AIE} followed by an overview of the SME algorithm in Section \ref{sec:SMEAnalogy}. Finally, a methodology for iterative rationale construction is provided in Section \ref{sec:rationale}, and experimental results are reported in Section \ref{sec:experiments}.

\subsection{Artificial Intentionality Engine}
\label{sec:AIE}

\noindent A simulated system known as the Artificial Intentionality Engine (AIE) has been developed. Within the AIE are a set of objects with various properties. The Agent has access to a set of actions which allows for travel between objects, transportation of objects and transformation of objects. Note that transformations may require the use of other colocated objects which operate as tools. Using reinforcement learning, an agent can develop policies for accomplishing various tasks. By observing a set of instances of each policy, the agent must first learn how to recognize instances of these behaviors. Taking the intentional stance, we then argue that the agent must extract a set of concepts that can be used to describe the free-floating rationale behind these behaviors. 

In terms of initial representation, we show how representative chronologies can be extracted and subsequently used to recognize novel instances of a behavior. A more conceptual understanding will be achieved via analogy to a library of prior experiences, where the rationale is already understood. A synopsis of the first five policies that have been constructed is given here. \textbf{Slumber} involves an agent transitioning from a waking to resting state due to fatigue. This transformation requires a bed to occur since it affords a place to rest. \textbf{Dinner} involves an agent consuming a chicken due to hunger. The concept of tools/utility is critical here since the agent must find a knife, transport the knife to the chicken and then dispatch it prior to consumption. \textbf{Chopping} involves an agent utilizing either a knife or axe to transform a tree into lumber (two distinct patterns are observed, one for each tool). The concept of danger is relevant since transporting knives often leads to injury. \textbf{Competition}, like consumption, requires an agent to use a knife to consume a chicken, however there exists an animal in the environment which will consume the chicken immediately if it observes the agent picking up the knife. The agent must then dispatch the animal prior to consuming the chicken. The rationale behind this behavior is similar to that of the cuckoo bird which must dispatch other entities in its nest to ensure its own survival. \textbf{Weather} requires the agent to discover the effects of the environment on its behavior. If the weather is good the agent may seek leisure, but if the weather is bad the agent must seek shelter.

\begin{figure}
    \centering
    \includegraphics[width=3.0in]{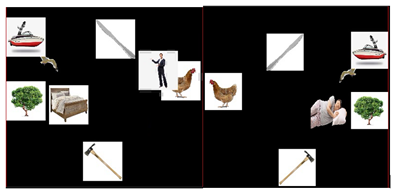}
    \caption{The AIE simulation associated with the slumber scenario.}
    \label{fig:AIE_slumber}
\end{figure}

\begin{figure}
    \centering
    \includegraphics[width=3.0in]{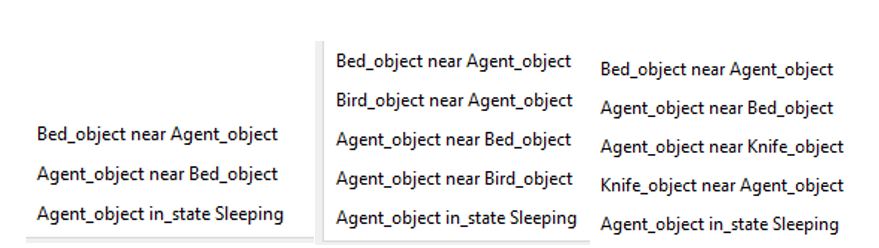}
    \caption{Three chronologies that describe the observed slumber behavior. The second and third chronology contain spurious events.}
    \label{fig:chronology}
\end{figure}

In the following example, the agent observes ten versions of the slumber behavior. Initial observations are a series of state specifications represented as a set of predicates. These predicates define i) the state of each object, ii) the proximity of each object to other objects, and iii) which objects are currently held by the agent. An initial state specification is randomly generated, and the state of the environment at the conclusion of each action is also made available to the agent (see Figure \ref{fig:AIE_slumber}). 
Three chronologies were constructed for this example, where a chronology embodies predicates that have changed from negative to positive. Each of the ten observations can be used to construct a chronology, and if multiple chronologies are sufficiently similar, a single representative chronology is selected. In this example three unique chronologies are automatically produced and used to represent the behavior. Note that all but the first chronology contains spurious events (see Figure \ref{fig:chronology}).

This process was repeated for all five behaviors. In order to gauge the classification capability of the representative chronologies, a distance metric was constructed which measures the degree to which a given chronology is consistent with an observed behavior. Using the average distance for two unseen observations of each behavior, a confusion matrix based on the distance measures is computed (see Table \ref{tab:chronology_accuracy}). While the diagonals are generally low (implying that novel instances can be accurately associated with their true behavior class), there are various off-diagonal entries with low distance measures. This implies that the chronology representation can be ambiguous. We thus argue for a more conceptually meaningful representation based on analogies with prior experiences. 


\begin{table}
\caption{Distance measures between estimated behavior chronologies and novel observations of such behaviors.}
{\scriptsize
\begin{center}
\begin{tabular}{l c c c c c}
    \hline
    Chronology & Slumber & Dinner & Chopping & Competition & Weather \\
    \hline
    Slumber & 0.00 & 1.00 & 1.00 & 1.00 & 1.00 \\
    Dinner & 0.67 & 0.10 & 0.40 & 0.31 & 0.50 \\
    Chopping & 1.00 & 0.60 & 0.00 & 0.62 & 0.00 \\
    Competition & 1.00 & 0.20 & 0.60 & 0.15 & 0.50 \\
    Weather & 0.67 & 0.90 & 0.80 & 0.93 & 0.00 \\
    \hline
\label{tab:chronology_accuracy}
\end{tabular}
\end{center}
}
\vspace{-10mm}
\end{table}

\subsection{Analogy Building under the Intentional Stance} 
\label{sec:SMEAnalogy}

\noindent Hofstadter \cite{Hofstadter} considers analogy to be a critical component of human reasoning, allowing us to make a ``mental leap’’ in order to understand and respond creatively to novel situations \cite{holyoak1996mental}. Inherent to analogy is a mapping, which is used to connect a \textit{base} domain – typically already known and familiar – to a \textit{target} domain, which is novel or less well understood. Structural Mapping Theory (SMT) \cite{gentner1983structure} provides a theoretical account of analogical mapping between a base and target and has been computationally
realized as the Structural Mapping Engine (SME) \cite{falkenhainer1989structure}.

SMT posits that analogies should be structural, which allows the objects in the base and target to have dissimilar surface-level properties. For example, the metaphor “but soft, what light through yonder window breaks? It is the east, and Juliet is the sun” \cite{shakespeare_mowat_werstine_2011} represents an attribute level comparison instead of a structural comparison and thus, under SMT they would not be mapped.
\textit{Attributes} can be represented in Predicate Calculus using unary predicates, e.g., $fairAtt(\textbf{sun})$ and $fairAtt(\textbf{Juliette})$ where $fairAtt(.)$ represents the object's attribute of beauty. 
\textit{Functions} also take one argument, but instead of being truth valued, they
range over symbols representing objects or quantities. For example, the mass of the Sun: $mass(\textbf{sun})$. Under SMT, since objects can be mapped, functional expressions can also be mapped. 
Binary and higher arity predicates represent \textit{relations} among objects,
and SMT allows their mapping under certain conditions, including that the relation name must be identical.
A classical analogy example from science is Rutherford’s development of the model of the atom (target), which took inspiration from the model of the solar system (base). The dynamics of the solar system were understood: the mass of the sun 
is greater than the mass of a planet and the sun attracts the planet which causes the planet to revolve around the sun. Structural similarities map the atom nucleus to the sun and electrons to planets. The nucleus attracts the electrons like the sun attracts the planets, and so on. Furthermore, this analogy allows for new potential inferences about the relationships between observations in the target domain that are transferred from the base domain. In this case, once the mapping is made, the observed mass differential and attraction can be inferred to be the cause of electrons revolving around the atom nucleus.

More formally, SME creates a map between a chosen base experience in a library of understood experiences $b_i \in B$ and novel target experience $t$ in four steps: local matching, global matching, hypothesis generation, and structural evaluation. 
The SME algorithm as described in \cite{falkenhainer1989structure} works by first finding all consistent pairwise correspondences between objects to create local match hypotheses $MH_{local}$s. Allowable object mappings are defined by a set of match rules. Here we start with traditional match rules of analogy in SME (see \cite{falkenhainer1989structure}, Appendix A) which ignore properties or attributes of objects, but identically match relations such as $greaterThan$ or $cause$ and extend this to flexibly match predicates within a category (e.g. the category of ``affordance'' predicates). These local matches are then combined into maximally consistent global mappings (Gmaps). A Gmap forms the maximal combinations of $MH_{local}$s, subject to two structural constraints. First, they must be one-to-one where an element in the base can at most correspond with a single element in the target and viceversa. Additionally, supports must map: when a $MH_{local}$ is made, the arguments within each of the elements must also map. It is possible for a base and target to have multiple Gmaps. From each Gmap, a set of candidate inferences (we will call them \textit{hypotheses} to emphasize that they are not logical consequences and may even be inconsistent with other statements in the target) that is derived. Hypotheses are new propositions constructed from the base that have not matching expression in the target but are structurally consistent with the rest of the relations in the current Gmap. In the last step, structural evaluation, evidence is attached to each match to score a Gmap. Individual $MH_{local}$s are given weights and these scores are summed across all matches within the Gmap. Better mappings, and thus better analogies, exhibit larger degrees of higher order relations. The best Gmap is then presented as a hypothesis $H$, the proposed analogy map.

In the past, the SME has been used to perform analogical reasoning between different domains, as illustrated by the science example of Rutherford's development of the model of the atom. There are also many examples (see \cite{falkenhainer1989structure}) where it is used as a means to understand the underlying causal mechanism of phenomena, for example understanding heat flow by analogy with water flow.
In this work, adopting the intentional stand, we are interested in explaining a sequence of observed events produced by one or more agents by hypothesizing the rationales behind the agents' actions.
We argue that this analogical process can lend itself to a deeper understanding of behaviors that we observe in others, allowing us to go beyond surface matching of the observed chronologies, as discussed in Sec.~\ref{sec:AIE}, to inferring rationales. 
We have re-purposed an SME to take in scenarios generated by observing an agent(s)
acting in a simulated or real world. This lets us integrate analogical reasoning with the AIE. Using a library of past experiences for which we understand the rationales, we can then perform analogical reasoning to hypothesize rationales and understand  novel experiences.

For the slumber example generated by the AIE, we can make an analogy to the rationale behind behavior of an agent when feeling cold. We assume that the agent has an existing experience for the rationale behind traveling to a bonfire when cold: when the agent travels to the fire she then becomes comfortable and knows the reason is because the fire affords properties such as warmth and brightness and the agent is cold (an unfulfilled desire). When this prior experience maps to the new chronology of slumber generated by the AIE, the $\textbf{bed}$ maps to $\textbf{fire}$; $asleepTf(.)$, a transformation, maps to the predicate of $comfortableTf(.)$; $tiredDes(.)$, an unfulfilled desire, maps to $coldDes(.)$; and affordances such as $flatAff(.)$ and $softAff(.)$ map to $warmAff(.)$ and $brightAff(.)$. From this, new inferences can be made connecting the unfulfilled desires to affordances as well as rationale behind observed behaviors. 
By adopting the intentional stance, the AIE describes experiences in terms of 
affordances, desires and transformations.
We modified the original SME formulation to 
flexibly capture comparisons among these types of relations, 
as in the mapping between $flatAff(.)$ and $warmAff(.)$. Our matching rules still 
require that matched predicates belong to the same category (e.g. both are affordances), and that predicate name matching occurs both for relations and for attributes. This extension is well suited to mapping rationales because mapped analogies with the same structure may be indicative of an overarching narrative, which can be understood by looking at category level structure. For example, in this case, the category level narrative might be: an agent goes to an object and then its state is transformed. This is because the agent had an unfulfilled desire and that object had affordances which could fulfill the desire. 

\begin{figure}
    \centering
    \includegraphics[width=3.0in]{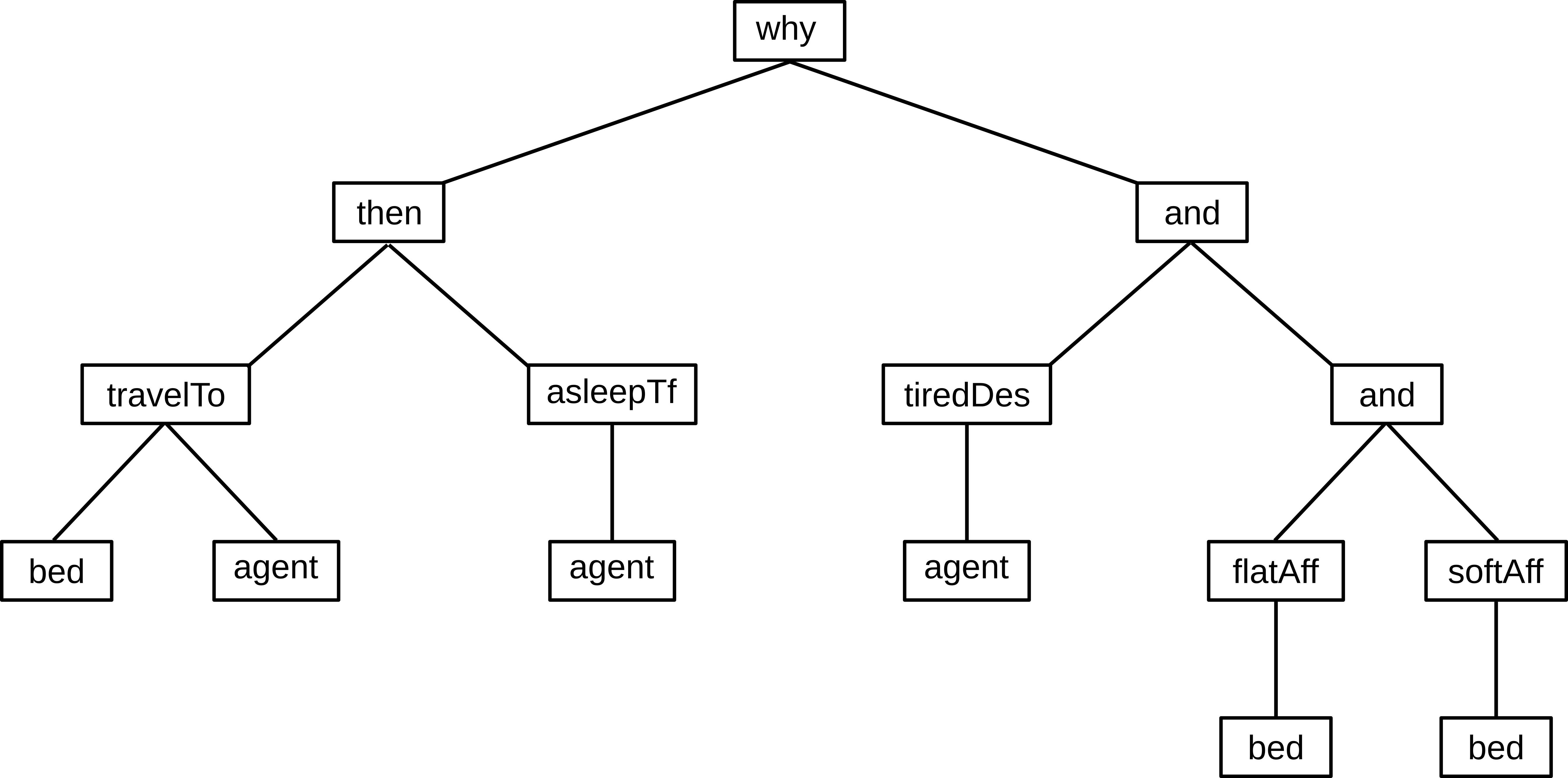}
    \caption{Resulting mapped analogy in target domain (tired) from base experience (cold) to make new inferences about the rationale $why(.,.)$ behind observed behavior.}
    \label{fig:slumberMap}
\end{figure}

Similarly, for the chopping example, given known information in our base about the ability of a hammer and rock to both pound in a nail, we can make a full analogy to chopping wood using an axe versus a knife. Here, $\textbf{axe}$ is mapped to $\textbf{hammer}$ and $\textbf{knife}$ is mapped to $\textbf{rock}$. The key aspect of this analogy is that, while both tools achieve the desired effect -- transforming $poundedTf(\textbf{nail})$ and $choppedTf(\textbf{wood})$, there is an advantage (represented as the relation $advantage(.,.)$ to one choice over the other in terms of a property of the items, $forceFn(.)$ for the hammer and $safetyFn(.)$ for the axe. We are also able to infer the causal nature of the relationships and elements of the rationale, from the mapping.

\begin{figure}
    \centering
    \includegraphics[width=3.0in]{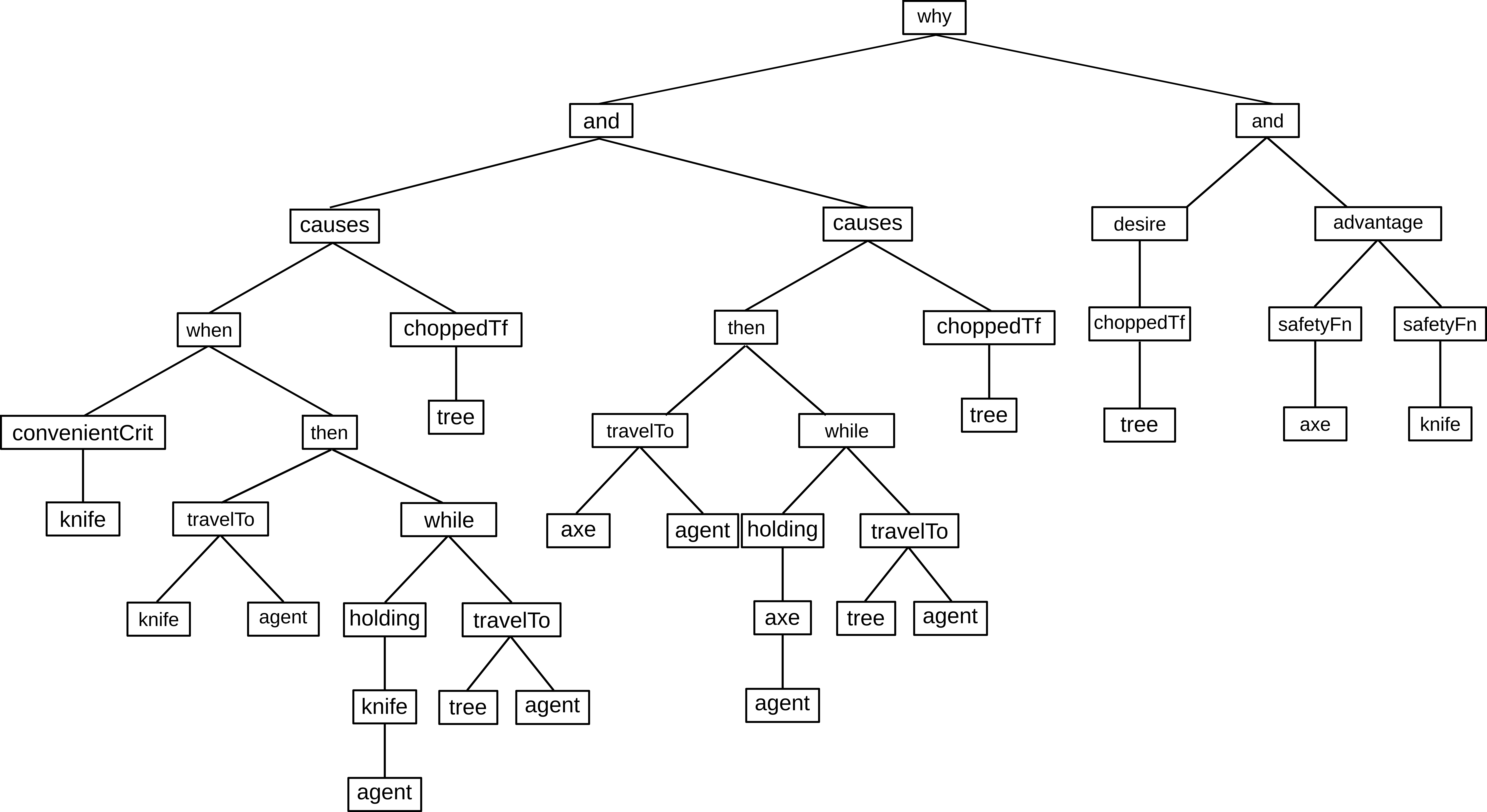}
    \caption{Rationale from analogy between chopping wood and pounding a nail.}
    \label{fig:choppingMap}
\end{figure}

\subsection{Iterative Rationale Construction}
\label{sec:rationale}

\noindent When considering observed novel experiences, except for the simplest cases, they will often require multiple analogies with past experiences to construct a rationale that fully explains the observations. 
The objective is then to hypothesize a rationale for a new experience
by collecting hypotheses drawn by analogy with multiple past experiences which together form a rationale for the novel experience. 
This objective drives the idea of analogy synthesis by combining partial hypothesized rationales distributed over different experiences via sequential pairwise analogy evaluation. As described in Section~\ref{sec:SMEAnalogy}, the SME constructs a mapping between a base $b$---a prior experience---and the target $t$---the novel situation. Given a set of bases $B =\{ b_1, b_2, \dots\ b_n \}$, corresponding to $n$ past experiences,  $b_1\dots b_n$, the idea is to attempt a pairwise mapping $b_i \rightarrow t$ for each $b_i \in B$. From each pairwise mapping, as previously explained, a set of hypotheses $h$ is derived that represent candidate rationales for transfer from the base $b_i$ to the target $t$. When $h \neq \varnothing$, the proposed algorithm augments the target $t$ with the rationale hypothesized by the analogy: $t \coloneqq t \bigcup h$. 
If $h$ is empty for all $b_i$, i.e., all $\{ b_1, b_2, \dots\ b_n \}$ have been paired with the current target $t$ and no new hypotheses were generated, the algorithm terminates.

\begin{algorithm}
\caption{Analogy Synthesis Algorithm}\label{alg:analogy_synthesis}
\begin{algorithmic}
\Require $B \neq \varnothing$
\Require $t \neq \varnothing$
\State $i \gets 1$
\State $b \gets b_i$
\Repeat
\State $newHyp \gets False$
\For{$b_i \in B$}
\State{$h = SME_{mapping}(b_i \rightarrow t)$}
\If{$h \neq \varnothing $}
    \State $t \gets {t \bigcup h}$ \Comment{t gets augmented}
    \State $newHyp \gets True$
\EndIf
\EndFor
\Until{$newHyp = False$}
\end{algorithmic}
\end{algorithm}
 
The process of pairwise analogical mapping results in incremental augmentation of the target $t$ with hypothesized rationales that partially explain the observations, thereby enabling an iterative full rationale construction. 
We have observed that the order in which past experiences $b_i$ are considered  matters. The final results may differ if the existence of an object is hypothesized (and a new symbol is introduced to represent the object) without matching it to an object explicitly mentioned in the target experience. The order matters because 
analogy with another base may find a matching object, which would make the introduction of a new symbol unnecessary. In cases where none of the pairwise analogies introduces new objects, the results will be the same regardless of the order.
Nevertheless, the order in which the bases are considered still matters in terms of computation time, i.e., it affects the number of iterations the algorithm goes through the entire set $B$ of bases before it terminates. 
We evaluated heuristics that utilize both structural as well as other attributes between the bases in $B$ and the target $t$ to establish some metrics we can use 
to choose an ordering of the bases. Both $t$ as well as each past experience $b_i$ contain hypothesized rationales expressed via structurally connected predicates and objects. As we have seen, the predicates contained in both $b_i$ and $t$ belong to 
different categories.
Few examples to illustrate the concept:\footnote{The SME uses Lisp notation for expressions, so from here on we use that notation.} in expression $(\it{travelTo} \bold{~Door} \bold{~Customer})$, the binary predicate $\it{travelTo}$ is of type $\it{relation}$. Predicate $\it{dangerAff}$, as in $(\it{dangerAff} \bold{~Weapon})$, is of type
$\it{affordance}$, and $(heightFn\bold{~Building})$ has predicate $heightFn$ of type $\it{function}$. The SME treats each type of predicate differently. When matching relations, the predicate name and the arguments must match. Affordances, functions, and desires, can match if they are of the same type even if the predicate names are different, e.g. $heightFn$ and $temperatureFn$, as long as the arguments match. 
The overall structural connectivity between predicates and objects is captured by a graph that is composed of nodes and edges.


The heuristic method for deciding the order in which to process the past experiences is based on combining a) pairwise similarity of predicates between the target $\it{t}$ and each $b_i \in{B}$ and b) on the structural richness of each $\it{b_i}$, assessed on the basis of the number of edges contained in the graph of each past experience $\it{b_i}$. We argue that a given past experience $\it{b_i}$ with higher structural complexity, i.e., higher number of edges, 
is more likely to generate hypotheses than simpler ones.
We further argue that weighting each $\it{b_i}$ based on the degree of similarity of predicates to the target $t$, further promotes past experiences that are more relevant to the novel target $\it{t}$. The method described is summarized below. 

\begin{algorithm}
\caption{Heuristic on Sequencing Partial Analogies}\label{alg:heuristic}
\begin{algorithmic}
\State{$ S_{b_i} \gets Compute ~predicate ~similarity ~to ~$t$, ~\forall ~b_i \in B$}
\State{$ Edges_{b_i} \gets Compute ~number ~of ~edges, ~\forall ~b_i \in B$}
\State{$ W_{b_i} \gets S_{b_i} \cdot Edges_{b_i}$}
\State{$ Sort ~{b_i} ~based ~on ~W_{b_i}$}
\end{algorithmic}
\end{algorithm}

\section{Experiments}
\label{sec:experiments}

\noindent To explore the plausibility of the approach, we have implemented Algorithm~\ref{alg:analogy_synthesis} 
and tested it on a representation of the gas pump experiences example described in the introduction. Our experiments involves four base experiences:

\begin{enumerate}
    \item Normal gas station visit experience: a customer travels to the gas stations, pumps gas, pays for the gas
    and leaves;
    \item Dog chases a person: an aggressive dog chases a person and the person runs away;
    \item Dark alley: a person walks into a dark alley, another person walks toward the first person and then attacks the latter;
    \item Car fire: a person travels to the gas station to pump gas, the car then catches fire and the person flees from
    the scene.
\end{enumerate}

The representation of the four base experiences is shown in Figures~\ref{fig:base_gas_station_visit}-\ref{fig:base_car_fire}, respectively.
Each base includes some statements about object attributes and relationships, e.g., that the dog is aggressive or that the gas station is not a socializing area. It also includes a number of events, similar to the chronologies generated by the AIE, e.g.,
that the customer traveled to the gas station. Finally, a rationale, represented by a proposition using the $why$ relation, is included as well. In each base, object names include a postfix to make sure they are unique, e.g., the object 
$gas\_station\_gsMt$ from the normal gas station ``micro-theory'' includes the postfix $\_gsMt$ to make it unique.

\begin{figure}
    \fbox{%
	    \parbox{0.95\linewidth}
	        {%
                 \scriptsize\it{
        	     (not\_social\_area\ \ gas\_station\_gsMt)\\
                (want\ \  gas\_gsMt\ \ customer\_gsMt)\\
                (sells\ \ gas\_gsMt\ \  gas\_station\_gsMt)\\
                \\
                (travelTo\ \  gas\_station\_gsMt customer\_gsMt)\\
                (pump\ \ gas\_gsMt\ \  customer\_gsMt)\\
                (pay\ \ gas\_gsMt\ \ customer\_gsMt)\\
                (travelTo\ \ somewhere\_gsMt\ \ customer)\\
                \\
                (why\\
                (and (travelTo\ \ gas\_station\_gsMt\ \ customer\_gsMt)\\
                \hspace*{2em} (pump\ \ gas\_gsMt\ \ customer\_gsMt))\\
                (and (want\ \ gas\_gsMt\ \ customer\_gsMt)\\
                \hspace*{2em} (sells\ \ gas\_gsMt\ \ gas\_station\_gsMt)))
                 }
	        }%
        }
    \caption{Past Experience - Normal gas station visit}
    \label{fig:base_gas_station_visit}
\end{figure}

\begin{figure}
    \fbox{%
	    \parbox{0.95\linewidth}
	        {%
                \scriptsize\it{    	
        	    (safeDesire person\_dcMt)\\
                (aggresive dog\_dcMt)\\
                \\
                (implies (aggressive dog\_dcMt) (dangerAff dog\_dcMt))\\
                \\
                (travelTo person\_dcMt dog\_dcMt)\\
                (flee person\_dcMt)\\
                \\
                (why \\ (flee person\_dcMt)\\
                (and (dangerAff dog\_dcMt) (safeDesire person\_dcMt)))
                }
	        }%
        }
    \caption{Past Experience - Chased by aggressive dog}
    \label{fig:base_chased_by_dog}
\end{figure}

\begin{figure}
    \fbox{%
	    \parbox{0.95\linewidth}
	        {%
                \scriptsize\it{    	
        	    (stranger person1\_daMt)\\
                (not\_social\_area darkAlley)\\
                (criminalDesire person1\_daMt)\\
                \\
                (implies (and (stranger person1\_daMt) (not\_social\_area darkAlley))\\ (dangerAff person1\_daMt))\\
                \\
                (travelTo darkAlley person2\_daMt)\\
                (travelTo person2\_daMt person1\_daMt)\\
                (attack   person2\_daMt person1\_daMt)\\
                \\
                (why\\
                (and (travelTo person2\_daMt person1\_daMt)\\
                \hspace*{2em}    (attack   person2\_daMt person1\_daMt))\\
                (criminalDesire person1\_daMt))
                }
	        }%
        }
    \caption{Past Experience - Dark alley attack}
    \label{fig:base_dark_alley_attack}
\end{figure}

\begin{figure}
    \fbox{%
	    \parbox{0.95\linewidth}
	        {%
                \scriptsize\it{    	
        	    (not\_social\_area gas\_station\_cfMt)\\
                (safeDesire customer\_cfMt)\\
                (want gas\_cfMt customer\_cfMt)\\
                (sells gas\_cfMt gas\_station\_cfMt)\\
                \\
                (travelTo gas\_station\_cfMt customer\_cfMt)\\
                (pump gas\_cfMt customer\_cfMt)\\
                (catchFire car\_cfMt)\\
                (flee customer\_cfMt)\\
                \\
                (causes (catchFire car\_cfMt) (dangerAff car\_cfMt))\\
                \\
                (why \\(flee customer\_cfMt)\\
                (and (dangerAff car\_cfMt) (safeDesire customer\_cfMt)))
                }
	        }%
        }
    \caption{Past Experience - Car catches fire}
    \label{fig:base_car_fire}
\end{figure}

In the novel experience, a customer visits the gas station to pump gas, then a stranger walks toward the customer, and the customer flees the scene. 
The task is to use our iterative analogy approach to hypothesize what might have been the rationale behind the observed behavior.
The representation of the target (new experience) is shown in Figure~\ref{fig:target}.

\begin{figure}
    \fbox{%
	    \parbox{0.95\linewidth}
	        {%
                \scriptsize\it{    	
                (sells gas gas\_station)\\
        	    (not\_social\_area gas\_station)\\
                (safeDesire customer)\\
                (stranger person)\\
                \\
                (travelTo gas\_station customer)\\
                (pump gas customer)\\
                (travelTo customer person)\\
                (flee customer)
            }
	        }%
        }
    \caption{Novel Situation - target $t$}
    \label{fig:target}
\end{figure}

As in the case of the bases, the target includes some statements about the objects, for example, that the gas station sells gas and that the customer has a desire to be safe. Other statements correspond to the chronology of observed events: 
the customer goes to the gas station, pumps gas, then another person walks to the customer (for simplicity we use the {\it travelTo} predicate to represent all the actions of going from one place to another), and then the customer flees the scene.
Unlike the bases, this being the novel experience, it is missing a rationale, which
is what we want to compute.


Applying the heuristic method outlined in the previous section to our set of bases and target suggests the following ordering of the bases:

\begin{easylist}
  \ListProperties(Start1=1, Mark1={.}, Progressive*=3ex)
  @ Normal gas station visit
  @ Dark alley
  @ Dog chases a person
  @ Car fire
\end{easylist}

\begin{figure}
    \centering
    \includegraphics[width=1.0\linewidth]{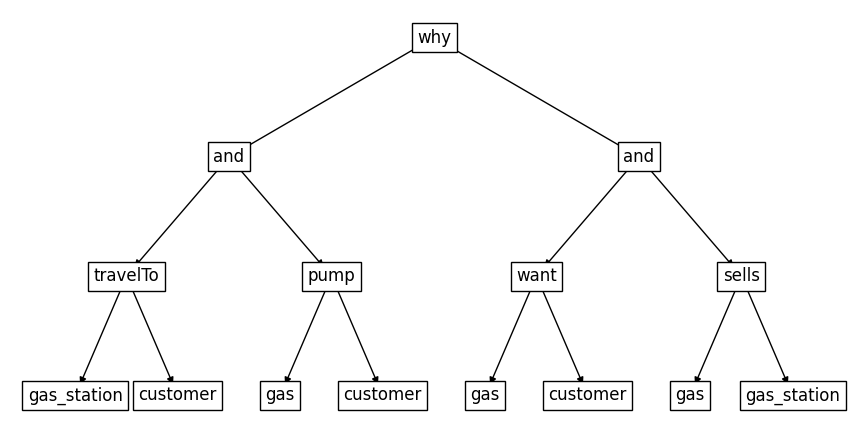}
    \caption{Hypotheses - Deduced from normal gas station visit}
    \label{fig:Hypotheses - Deduced from Normal Gas Station Visit}
\end{figure}

Next we apply Algorithm~\ref{alg:analogy_synthesis} modified to consider the bases in the order suggested by the heuristics. The results from the first base include four new hypotheses, one of them being a partial rationale, shown as a graph in Figure~\ref{fig:Hypotheses - Deduced from Normal Gas Station Visit}, for the target behavior. The hypothesis gives us only a rationale for the customer going to the gas station and pumping gas, so it is clearly partial as it does not explain the other observed behavior.

\begin{figure}
    \centering
    \includegraphics[width=0.6\linewidth]{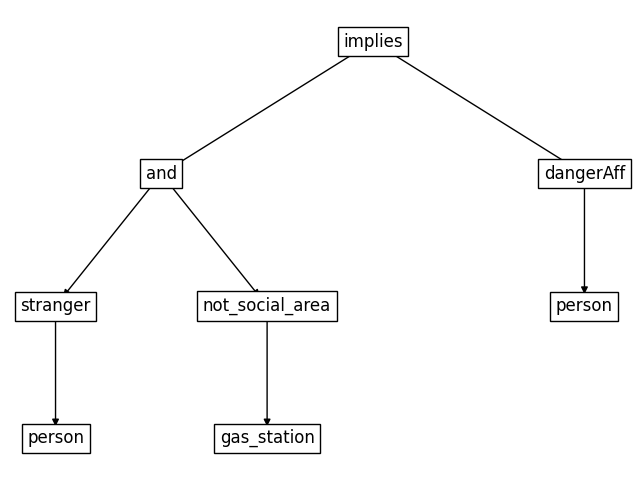}%
    \includegraphics[width=0.5\linewidth]{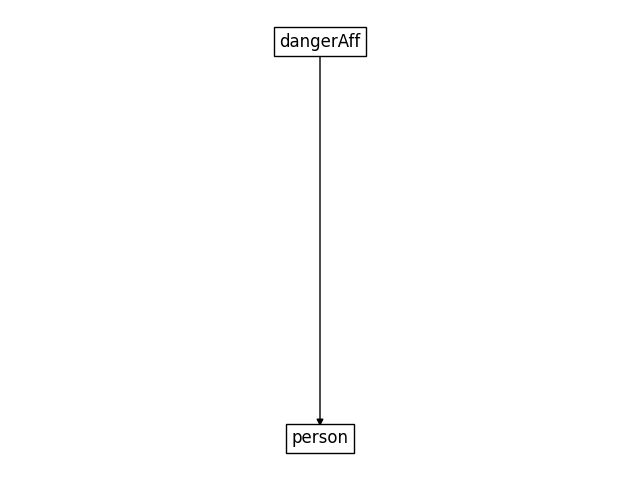}
    \caption{Hypotheses - Deduced from Dark Alley}
    \label{fig:Hypotheses - Deduced from Dark Alley}
\end{figure}

The algorithm then adds the four hypotheses to the target and considers the next base. From the dark alley base, we obtain seven new hypotheses, three of which are discarded because they involve unobserved events.\footnote{We assume that all actions are observable and the chronology part of the targets is complete. Hence hypotheses about unobserved events are discarded.} This time, we do not obtain a rationale, but do obtain an importance piece of information: that the stranger in the non-social gas station is dangerous. These hypotheses are depicted in 
Figure~\ref{fig:Hypotheses - Deduced from Dark Alley}.

\begin{figure}
    \centering
    \includegraphics[width=.5\linewidth]{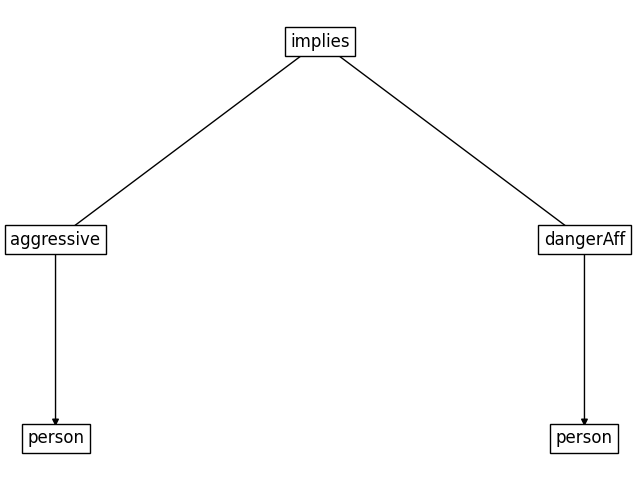}%
    \includegraphics[width=.5\linewidth]{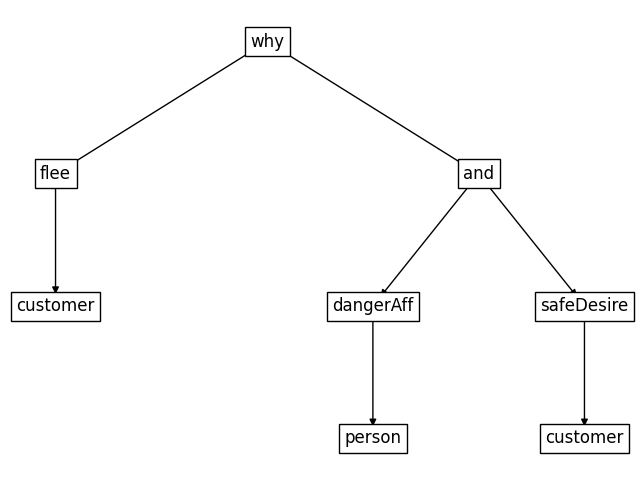}
    \caption{Hypotheses - Deduced from Chased by Dog}
    \label{fig:Hypotheses - Deduced from Chased by Dog}
\end{figure}

Then next base, the dog chase experience, results in five new hypotheses, including a rationale for the fleeing behavior and another hypothesis around the danger affordance of the person---that the person is aggressive. 
The hypotheses are depicted in 
Figure~\ref{fig:Hypotheses - Deduced from Chased by Dog}.

At this point, we have collected rationales for all the events in the target and two hypotheses that explain why the person that walked to the customer has the danger affordance: because he is a stranger or because he is aggressive. The algorithm will next consider the last remaining base, the car fire case. As expected, given all the hypotheses we have collected so far this based does not yield any further hypotheses. The algorithm will then do another iteration over the set of bases, which will generate no further hypotheses and will terminate.

\begin{figure}
    \centering
    \includegraphics[width=0.6\linewidth]{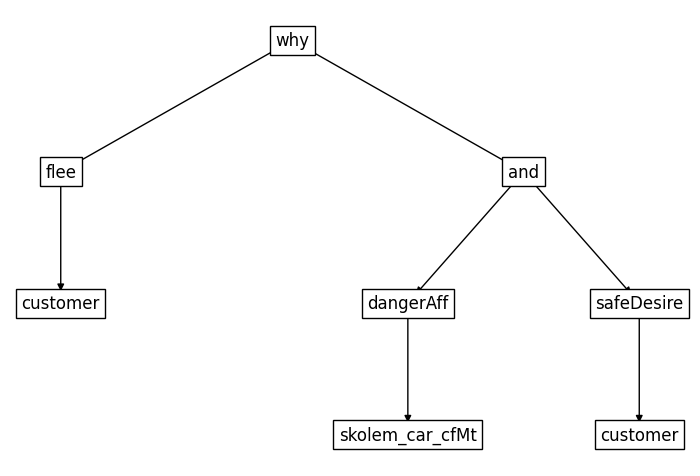}%
    \includegraphics[width=0.5\linewidth]{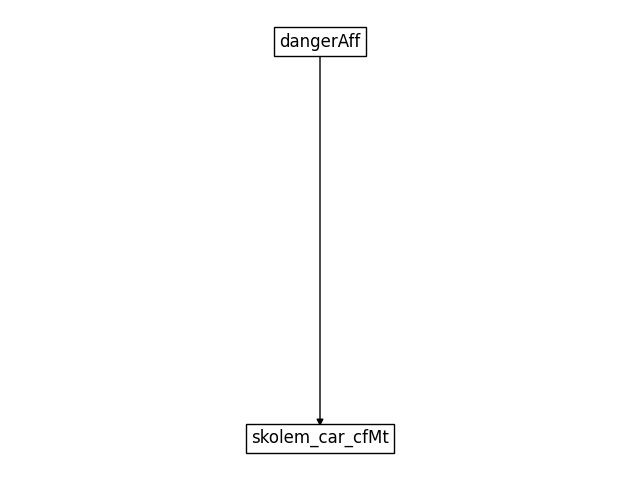}
    \caption{Hypotheses - Deduced from Car on Fire with Skolem Symbol}
    \label{fig:Hypotheses - Deduced from Car on Fire with Skolem Symbol}
\end{figure}

To illustrate how the results may differ if the bases are considered in a different order, let us look at the results of matching the car fire base with the original target, i.e., before any other base is considered. The result is five hypotheses.
Interestingly, the car fire base does not contain expressions that can lead to matching any base object with the stranger person in the target. Instead, the analogy hypothesizes the existence of an object that catches fire and becomes dangerous. The conjectured object is new (does not appear in the target) and is introduced via a skolem constant $skolem\_car\_cfMt$. The symbol includes the postfix $car\_cfMt$ to indicate its origin but that does not influence how it is treated by the system. 
The matching results in six hypotheses, two of which are discarded. The remaining for hypotheses include a rationale for the fleeing event, shown in
Figure~\ref{fig:Hypotheses - Deduced from Car on Fire with Skolem Symbol}.

Intuitively, the hypotheses say that the rationale behind the customer fleeing is that there is an unknown dangerous object (represented by the new skolem constant) and the customer has a desire to be safe.

By following the ordering of the bases suggested by the heuristics, the need to introduce new object symbols is avoided as bases that are more likely to generate more matches are given preference.

\section{Conclusion}

\noindent In order to understand and formalize rationale, we have introduced the components of a novel modeling pipeline. First, the AIE can be leveraged to generate behaviors and construct chronologies of events, then SME provides a mechanism to map analogies between previous experiences and new observations, and finally novel synthesis techniques allow this mapping to form flexible, partial analogies.

We argue that a given analogy may produce various hypotheses which can all potentially be valid. For example when considering the prior experience ``man runs away from fire,'' one might produce the hypothesis that people run because they are afraid of a dangerous thing. Another prior experience might be ``man starts running when the pistol of the race fires.'' In this case it is not fear that causes the person to run, it is the desire to win the race or maybe increase ones fitness. So we argue that both these hypotheses could be correct. Our algorithm for analogy synthesis provides a partial mapping strategy designed to generate these plausible explanations through comparison to existing experiences. Having said that, as part of future step in our effort is to study the mechanism to validate the hypotheses generated, and autonomously reason over the most appropriate hypothesis that best explains the observation of the current novel situation.

{\small
\bibliographystyle{ieee}
\bibliography{GEbib}
}

\end{document}